\documentclass{article}

     \usepackage[preprint, nonatbib]{neurips_2019}

\usepackage[utf8]{inputenc} % allow utf-8 input
\usepackage[T1]{fontenc}    % use 8-bit T1 fonts
\usepackage{url}            % simple URL typesetting
\usepackage{booktabs}       % professional-quality tables
\usepackage{amsfonts}       % blackboard math symbols
\usepackage{nicefrac}       % compact symbols for 1/2, etc.
\usepackage{microtype}      % microtypography
\usepackage{graphicx}
\usepackage{physics}
\usepackage{appendix}

\title{GAS-NeXt: Few-Shot Cross-Lingual Font Generator}

\author{%
  Haoyang He$^*$\\
  Electrical and Computer Engineering\\
  Carnegie Mellon University\\
  Pittsburgh, PA 15213 \\
  \texttt{hehaoyang@cmu.edu} \\
  \And
  Xin Jin$^*$\\
  Electrical and Computer Engineering\\
  Carnegie Mellon University\\
  Pittsburgh, PA 15213 \\
  \texttt{xinj2@andrew.cmu.edu} \\
  \And
  Angela Chen$^*$\\
  Electrical and Computer Engineering\\
  Carnegie Mellon University\\
  Pittsburgh, PA 15213 \\
  \texttt{xinyuc2@andrew.cmu.edu} \\
}

\begin{document}

\maketitle

%\begin{abstract}

%\end{abstract}

\def\thefootnote{*}\footnotetext{Authors contributed equally to this work}\def\thefootnote{\arabic{footnote}}
\def\thefootnote{}\footnotetext{Our code can be found at \url{https://github.com/cmu-11785-F22-55/GAS-NeXt}}\def\thefootnote{\arabic{footnote}}

\begin{abstract}
Generating new fonts is a time-consuming and labor-intensive task, especially in a language with a huge amount of characters like Chinese. Various deep learning models have demonstrated the ability to efficiently generate new fonts with a few reference characters of that style, but few models support cross-lingual font generation. This paper presents GAS-NeXt, a novel few-shot cross-lingual font generator based on AGIS-Net and Font Translator GAN, and improve the performance metrics such as Fréchet Inception Distance (FID), Structural Similarity Index Measure(SSIM), and Pixel-level Accuracy (pix-acc). Our approaches include replacing the original encoder and decoder with the idea of layer attention and context-aware attention from Font Translator GAN, while utilizing the shape, texture, and local discriminators of AGIS-Net. In our experiment on English-to-Chinese font translation, we observed better results in fonts with distinct local features than conventional Chinese fonts compared to results obtained from Font Translator GAN. We also validate our method on multiple languages and datasets.
\end{abstract}

\section{Introduction}
Generating new fonts is a time-consuming and labor-intensive task, especially in a language with a huge amount of characters like Chinese. Various deep learning models have demonstrated the ability to efficiently generate new fonts with a few reference characters of that style, but few models support cross-lingual font generation. This paper aims to develop a few-shot cross-lingual font generator based on AGIS-Net and Font Translator GAN and improve the performance metrics mentioned in Section \ref{section:baseline}. Our approaches include redesigning the encoder and adding a local discriminator. We will validate our method on multiple languages and datasets mentioned in Section \ref{section:dataset}.

\section{Related Work}

\subsection{Generative Adversarial Network}

Generative Adversarial Networks is a framework for estimating generative models via the adversarial net, first proposed by Goodfellow et al. in 2014 \cite{GAN}, featuring two neural networks (generator and discriminator) competing against each other for a zero-sum game to achieve an optimized solution. Since its introduction, there are several focuses that aim to improve the performance of the generative models. Mirza et al. introduced conditional GAN that allows the addition of conditions on both the generator and discriminator in GANs \cite{cGAN}. 
% Radford et al. introduced an unsupervised representational learning method for deep convolutional GANs \cite{radford2015}. Salimans et al. proposed an improved training technique for semi-supervised GANs to generate images that humans find more realistic \cite{salimans2016}. 
% Chen et al. proposed InfoGAN that maximizes the mutual information between a small subset of latent variables and the observations and learns disentangled information unsupervised \cite{infoGAN}. 
% Odena et al. introduced a method that employs label conditioning in image synthesis to produce photorealistic generated images with GANs \cite{odena2017}. 
Isola et al. proposed $Pix2Pix$, an image-to-image translation network based on GANs \cite{pix2pix}. Zhu et al. refined the image-to-image translation problem by introducing a multimodal variant BicycleGAN and modeling a distribution of possible outputs in conditional GAN setting \cite{zhu2017}. 
% Miao et al. proposed a method to encourage the generator to explore minor modes with a regularization method to solve the mode collapse problem for conditional GANs \cite{miao2019}. 
To solve the problem that paired data are hard to obtain, Lee et al. proposed a method for using unpaired data with disentangled representations \cite{DRIT}, and Zhu et al. proposed Cycle-consistent GANs to use learn a mapping such that the distribution of translated images is indistinguishable from the target image distribution using an adversarial loss \cite{cyclegan_zhu}.

As the application of image generation is usually in situations where abundant training data is not available, more recent approaches have been focused on few-shot learning-based image generation tasks. Clouâtre and Demers proposed FIGR that mediates the lack of training data with a GAN meta-trained by Reptile \cite{figr}. Liu et al. proposed FUNIT, an unsupervised few-shot image-to-image translation framework \cite{funit}.
% However, these methods have a problem in that the generated images have trouble retaining the structure or orientation of the original. Saito et al. proposed COCO-FUNIT that computes the style embedding of the example images conditioned on the input image \cite{coco-funit}.

\subsection{Font generation}

Font generation is a well-studied field. 
% Campbell and Kautz introduced the manifold of alphabetic font and learned new fonts by using the existing fonts for interpolation \cite{interpolation}. 
Yang et al. introduced a Generative Adversarial Network (GAN)-based approach in text style transfer \cite{Yang2019TETGAN}. Hayashi et al. proposed a GAN-based method to generate new fonts from seen fonts while maintaining consistency among the generated fonts and diversity from seen fonts \cite{GlyphGAN}. More recently, studies in font generation have been focused on non-alphabetic languages such as Chinese, as the number of classes of characters is huge compared to alphabetic languages, thus a font generator provides feasibility in applications. Tian proposed a conditional adversarial network for Chinese font generation with one-hot encoding for font styles, but its application is limited to seen font styles \cite{zi2zi}. Zhang et al. proposed a method of separating style and content in font generation \cite{zhang2018separating}. Gao et al. proposed a GAN that encodes content and a few samples of the style and decodes to the new style based on shape and texture and used a local discriminator in addition to the shape and texture discriminator \cite{AGIS}. 
% Chen et al. introduced a few-shot Chinese font generation method with deep meta-learning that utilizes the characteristics of existing fonts to avoid overfitting \cite{MLFont}. Jiang et al. proposed a one-shot Chinese character generation framework \cite{W-Net}. 
One of the latest approaches is Font Translator GAN (Font Translator GAN) \cite{ftransgan}. This method features cross-lingual font generation capability using layer attention and context-aware attention in its encoder and decoder. It incorporates the shape discriminator and texture discriminator but does not have a discriminator focusing on differentiating local features of the generated character. 

% \subsection{Component-based font generation}

One common approach to font generation is decomposing characters into smaller components and studying their localized features. This approach often results in better performance on characters with complex structures, such as Chinese characters which can be decomposed into approximately 500 radicals or 40 strokes. DM-Font \cite{DMFont} introduced the component-wise extractor and stored component features in dual memory. However, the model was only tested on simple language (Korean and Thai) with fixed component size and failed on language with complex glyphs such as Chinese characters. 
% LF-Font \cite{LFFont} tackled the problem by proposing low-rank factorization modules to generate component and style factors.
RD-GAN \cite{RDGAN} and CC-Font \cite{CCFont} both applied radical decomposition and succeeded in predicting characters of unseen components, but the former failed in the cross-lingual system. Calli-GAN \cite{CalliGAN} encoded the component sequence using a recurrent neural network and does not require component labels, but it failed on cross-lingual prediction or unseen styles and components. To generalize the model to a cross-lingual system, MX-Font \cite{MXFont} applied component-conditioned methods and interpreted the component assignment problem as a graph-matching problem. More recently, CG-GAN \cite{CGGAN} introduced the component-aware module with a simple generator that succeeded in the cross-lingual generation, and it can be further extended to scene text editing. 

% Decomposing handwritten characters is another challenging problem. Specifically, personalized Chinese font needs to generate at least 3,000 most commonly used characters but only with few training characters \cite{Chang_2018}. Previous works leverage CycleGAN \cite{Chang_2018, cyclegan_zhu} to automatically generate calligraphy work with aesthetic values. However, this method suffers the mode collapse issue. Designing an efficient and accurate decomposing strategy is quite crucial in order to compensate for information loss. Zeng et al. introduced one-bit stroke encoding and a stroke-encoding reconstruction loss imposed on the discriminator to preserve the key mode information of a Chinese character \cite{storkegan}. The stroke encoding successfully alleviates the issue in CycleGAN with improved preservation of strokes and diversity of generated character \cite{storkegan}.

\subsection{Style Transfer}
Style transfer is the process of adapting the style of an image to another source image to produce a new image with the adapted style while preserving the content of the source image. Gatys et al. first introduced a convolutional neural network-based method of image artistic style transfer \cite{gatys2016image}. 
% Johnson et al. furthered this approach by utilizing a feed-forward network for image translation and introduced a perpetual loss for training \cite{johnson2016perceptual}.
Huang and Belongie proposed a method that allows arbitrary style transfer in real-time by adaptive instance normalization \cite{huang2017arbitrary}. Li et al. added a pair of feature transforms, whitening, and coloring embedded in the image reconstruction network to improve the capability of generalizing to unseen styles and improve generation quality \cite{li2017universal}. 
% Kotovenko et al. proposed a method to reduce the amount of content removed from the source image by introducing a content transformation block between the encoder and the decoder \cite{kotovenko2019content}. 
Tuygen et al. introduced a method to generate multiple images with a single image-style pair by utilizing deep correlational multimodel style transfer \cite{tuyen2021deep}. The methods in image style transfer have also been explored in the style transfer applications in other forms of signals. Huang et al. proposed a method for video style transfer based on the feed-forward network for image style transfer \cite{huang2017real}. Verma and Smith adapted the work by Gatys et al. for audio style transfer \cite{verma2018neural}. 

\subsection{Similarity score}

Many of the generative models for images tend to produce blurry images with low bound approximation or restriction on the distribution \cite{Ghojogh_2020}. A popular loss like mean square error is not ideal for image quality assessment \cite{4775883}. Structural similarity Index (SSIM) is a perceptual measure shown to be perfect for different generative models and autoencoders. Using SSIM as the loss metric can improve the perceptual quality of the generated images \cite{4775883}.

\section{Baseline model}
\label{section:baseline}
We adopt the architecture of AGIS-Net\cite{AGIS} to be our baseline model as shown in Figure \ref{fig:baseline_overview} and evaluate its performance using Fr\'echet Inception Distance (FID), Structural Similarity Index Measure(SSIM), and Pixel-level Accuracy (pix-acc). 

\subsection{Problem Formulation}
Our goal is to generate a stylized glyph image with the provided content and a few sample images of the specified style. We need two sets of input: the content reference image $x_C$ and the style reference input $X_s$. The content reference image $x_C$ is a binary image in standard font with minimal style information. Given that the stylized examples are usually limited, we use $n$ to denote the size of the few-shot style reference set $R_s$, and we randomly sample $m \leq n$ images from the reference set as style reference input $X_s \subseteq R_s$.

\subsection{Baseline Selection}
Before AGIS-Net, previous approaches either focuses on shape synthesis or were designed for texture transfer. Campbell and Kautz \cite{interpolation} built a font manifold for Latin glyphs' outlines. However, it can fail when the complexity of glyphs greatly increases. Zeng et al. \cite{storkegan} attempted to utilize strokes from Chinese glyphs and learn to produce corresponding strokes for others but in the same style. However, it is not suitable for other scenarios like synthesizing Latin glyphs. Therefore, the most relevant work for this task is MC-GAN \cite{mcgan} which is the first architecture that solved learning styles of both texture and shape. Moreover, it is capable to do glyph shape synthesis and texture transfer as well. However, it is limited to 26 Latin capital letters as input, and it performs badly in generating other writing systems like Chinese characters. Thus, it's definitely not the best choice as the baseline model for our task. AGIS-Net \cite{AGIS} can effectively solve the above-mentioned problems. Compared to MC-GAN \cite{mcgan}, it has a wider application scenario. In other words, it can not only handle Latin glyphs but also Chinese and any other writing systems. Other competitors like BicycleGAN \cite{cyclegan_zhu} which is a leading general-purpose image-to-image translation model and TETGAN \cite{Yang2019TETGAN} which is the state-of-the-art artistic font style transfer model, perform slightly worse for cross-lingual tasks according to the performance result by Gao et al. \cite{AGIS}. Overall, we believe that AGIS-Net is a competitive and robust model for our task.

\subsection{Model Description}

\begin{figure}
    \centering
    \includegraphics[width=1.0\textwidth]{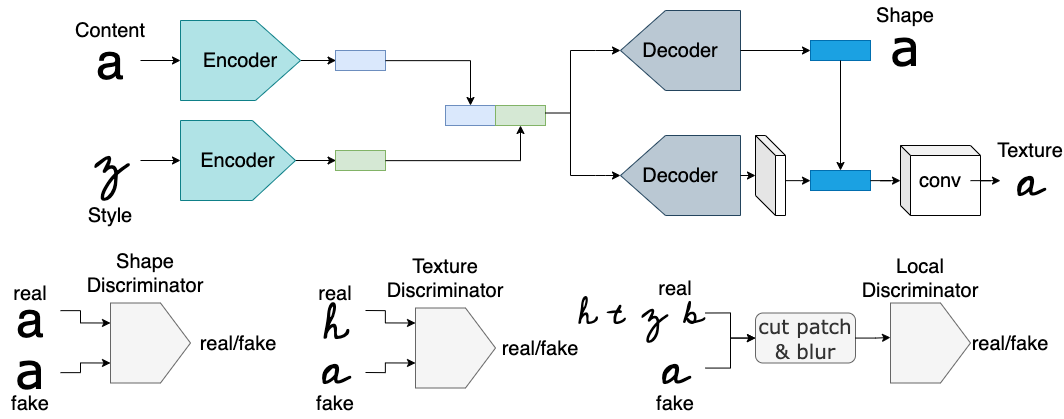}
    \caption{\label{fig:baseline_overview}AGIS-Net Architecture. The upper part is the generator consisting of two parallel-working encoders and decoders, and the lower three are discriminators.}
\end{figure}

%  Summary
% We choose the state-of-art one-stage few-shot cross-lingual font generation model AGIS-Net as our baseline. 
AGIS-Net consists of a generator and three discriminators, as shown in Fig. \ref{fig:baseline_overview}. The generator has two parallel working branches of encoders and decoders, where the encoders are responsible for extracting the content and style features independently, and the texture decoder generates the synthesized image collaboratively with the shape decoder. The output synthesized image is then fed into the shape and texture discriminators as negative samples during training, and the real glyph images are cut patched and blurred to introduce more fake samples for the local discriminator. The local discriminator helps generate more realistic images with reduced noise and alleviates the problem of unbalanced positive and negative glyph images in the few-shot setting.

%  Generator
\textbf{Generator:} Fig. \ref{fig:baseline_generator} shows the architecture of the generator. A binary image with minimal style information is inputted to the content encoder, and a set of stylized images are randomly sampled from the few-shot reference set as the input to the style encoder. The content and style features are extracted from several down-sampling convolution layers %TO DO: add details% 
and skipped connected to the input of corresponding decoder layers as shown in \ref{fig:baseline_submodule}. The shape decoder concatenates the output from the previous layer and the extracted content and style features from the corresponding encoder layer to generate the shape image $y_{gray}$. The texture decoder further combines the texture information with the shape information through concatenation, and the synthesized image $y$ is produced by feeding the concatenated shape image $y_{gray}$ and texture information to a convolution layer.

%  Objective Function
\textbf{Objective Function:} The objective function consists of four different loss functions: the adversarial loss, the $L_1$ loss, the contextual loss, and the local texture refinement loss
$$L(G, D_{sha}, D_{tex}, D_{local}) = L_{adv} + L_1 + L_{CX} + L_{local},$$
where $G$ is our generator, $D_{sha}$, $D_{tex}$, and $D_{local}$ represent the shape, texture, and local discriminator respectively. We train our model by optimizing
$$\min_G \max_{D_{sha},D_{tex},D_{local}} L(G, D_{sha}, D_{tex}, D_{local}).$$

\begin{itemize}
    \item \textbf{Adversarial Loss:}
    We train the generator $G$ and two discriminators $D_{sha}$ and $D_{tex}$ based on the following adversarial loss. $y_{gray}$ is the gray-scale image generated from the shape decoder, $y$ is the synthesized image, $t_s^r$ is the ground truth image, and $t_{gray}^s$ is the gray-scaled ground truth image used to train the shape discriminator. $\lambda_{adv\_sha}$ and $\lambda_{adv\_tex}$ are the weights set to 1.0 according to \cite{AGIS}.
    \begin{align*}
        L(D_{sha}) &= \mathbb{E}_{t_{gray}^r}[\log(D_{sha}(t_{gray}^r))] + \mathbb{E}_{y_{gray}}[\log(1-D_{sha}(y_{gray}))]\\
        L(D_{tex}) &= \mathbb{E}_{t_s^r}[\log(D_{tex}(t_s^r))] + \mathbb{E}_y[\log(1-D_{tex}(y))]\\
        L_{adv} &= \lambda_{adv\_sha} \mathbb{E}_{y_{gray}}[\log(1-D_{sha}(y_{gray}))] + \lambda_{adv\_tex} \mathbb{E}_y[\log(1-D_{tex}(y))]. \\
    \end{align*}
    
    \item \textbf{$L_1$ Loss:}
    $L_1$ Loss helps stabilize the training. We simply combine the $L_1$ loss of the gray-scale image and the texture image with weights $\lambda_{L_{1gray}}=50.0$ and $\lambda_{L_{1tex}}=100.0$. These weights are set to zero for the unseen data.
    $$L_{1} = \lambda_{L_{1gray}} \mathbb{E}_{y_{gray},\hat{y}_{gray}} \norm{y_{gray}-\hat{y}_{gray}}_1 + \lambda_{L_{1tex}} \mathbb{E}_{y,\hat{y}} \norm{y-\hat{y}}_1,$$
    where $\hat{y}$ and $\hat{y}_{gray}$ are ground truth images of $y$ and the gray-scale version of the ground truth, and we will reuse them in the following loss functions.
    
    \item \textbf{Contextual Loss:}
    Contextual loss measures the similarity between feature map collections of the images, and it focuses on high-level style features. The similarity $CX$ between $x_i$, the i-th feature map of image $X$, and $y_j$, the j-th feature map of image $Y$ is defined as the normalized exponential of the shifted normalized cosine distances
    $$ CX_{ij} = \frac{w_{ij}}{\sum_k w_{ik}}, \hat{d}_{ij} = \frac{d_{ij}}{\min_k d_{ik} + \epsilon},  w_{ij} = \exp{\frac{1-\hat{d_{ij}}}{h}}, $$
    $$CX(X,Y) = \frac{1}{N} \sum_j \max_i CX_{ij},$$
    where $d_{ij}$ is the cosine distance between $x_i$ and $y_j$, $h$ and $\epsilon$ are hyper-parameters set to 0.5 and $1e-5$ according to \cite{context_loss}. 
    
    We use pre-trained VGG19 to extract features. $\phi^l$ denotes the extracted features from the $l$-th layer of VGG19, and $L$ is the total number of layers. In our experiment, we use $\lambda_{CXgray}=15.0$ and $\lambda_{CXtex}=25.0$ for the weights. The contextual loss is formulated as
    
    \begin{align*}
        L_{CX} &= \lambda_{CX_{gray}} \Big(-\frac{1}{L} \sum_l \log(CX(\phi^l(y_{gray}),\phi^l(\hat{y}_{gray}))) \Big) \\
        &+ \lambda_{CX_{tex}} \Big(-\frac{1}{L} \sum_l \log(CX(\phi^l(y),\phi^l(\hat{y}))) \Big).\\
    \end{align*}
    
    \item \textbf{Local Textual Refinement Loss:}
    The local discriminator $D_{local}$ takes the cut patched real images $p_{real}$ as positive samples, the blurred real images  $p_{blur}$, and our synthesized patches $p_y$ as negative samples. We set the weights $\lambda_{local} = 1.0$ based on \cite{AGIS}.
    \begin{align*}
        L(D_{local}) & = \mathbb{E}_{p_{real}}[\log(D_{local}(p_{real}))]) + \mathbb{E}_{p_{blur}}[\log(1-D_{local}(p_{blur}))] + \\ &+ \mathbb{E}_{p_y}[log(1-D_{local}(p_y))],\\
        L_{local} &= \lambda_{local} \mathbb{E}_{p_y}[\log(1-D_{local}(p_y))].\\
    \end{align*}
\end{itemize}

% \begin{figure}
%     \centering
%     \includegraphics[width=\textwidth]{fig/baseline_model.png}
%     \caption{Baseline Model}
%     \label{fig:baseline_model}
% \end{figure}

\subsection{Evaluation Metric}
To evaluate the model performance, we adopt three commonly-used metrics in image generation tasks: Fr\'echet Inception Distance (FID), Structural Similarity Index (SSIM), and Pixel-level Accuracy (pix-acc).

\textbf{Fr\'echet Inception Distance (FID)}

FID compares the probability distributions of feature vectors. FID between the fake image distribution $x$ and the ground truth image distribution $y$ is computed as
$$FID = \norm{\mu_x - \mu_y}^2 + Tr(C_x+C_y-2\times \sqrt{C_x \times C_y}),$$
where $\mu_x$ and $C_x$ are the mean and covariance matrix of the feature vectors distribution $x$. We use the module "pytorch-fid" to generate our FID scores. 

\textbf{Structural Similarity Index (SSIM)}

SSIM measures the structural similarity between the synthesized image $x$ and the ground truth image $y$ by
\begin{align*}
    SSIM &= \frac{(2\mu_x \mu_y+c_1)(2\sigma_{xy}+c_2)}{(\mu_x^2+\mu_y^2+c_1)(\sigma_x^2+\sigma_y^2+c_2)}, \\
    % c_1 = (k_1)
\end{align*}
where $\mu_x$ and $\mu_y$ are the pixel sample mean of $x$ and $y$; $\sigma_x$, $\sigma_y$, and $\sigma_{xy}$ are the variance of $x$, $y$, and the covariance of $x$ and $y$; $c_1$ and $c_2$ are constants to avoid instability when the denominator is small. In our evaluation, we adopt "skimage.metrics" package to evaluate the SSIM score. 

\textbf{Pixel-level Accuracy (pix-acc)}
$$PA=\frac{\sum_{j=1}^{h}n_{jj}}{\sum_{j=1}^{k}t_j},$$
where $n_{jj}$ denotes the number of correctly classified pixels in class $j$, and $t_j$ is the number of pixels labelled as class $j$. We use the gray-scale image for the evaluation, and thus the class is binary. Pixel-level accuracy is then averaged to evaluate the overall quality of the synthesized image.

\section{Dataset}
\label{section:dataset}
Gao et al. proposed a large-scale dataset using professional-designed fonts in different shapes and texture styles\cite{AGIS}. We will use it to validate if our method is effective and extendable at the first stage. CASIA also provides standard datasets for handwritten Chinese character recognition\cite{casia}. The data has been annotated at the character level, so it's preferable to be tested on for evaluating performance metrics. For English glyph images, we use the dataset proposed by Azadi et al. \cite{azadi}. Moreover, regarding plain context inputs, we will take up the average font style as common as possible for Chinese characters and Code New Roman as MC-GAN\cite{mcgan} for English letters. For our extended model, we use the dataset proposed by Li et al. \cite{ftransgan}. This dataset is mainly for cross-lingual font style transfer. All the style inputs are gray-scale Latin fonts, and all the context inputs are Chinese characters. They are used to generate styled Chinese characters.

\section{Baseline Implementation Completeness}
\label{section:bic}

\begin{table}
\caption{\label{table:eval} Evaluation Metrics of AGIS-Net and our baseline model}
\begin{tabular}{|p{3cm}||p{3cm}|p{3cm}|p{3cm}|}
    \hline
    \multicolumn{4}{|c|}{Evaluation Metrics} \\
    \hline
     Model & FID & SSIM & pix-acc \\
     AGIS-Net & 74.567 & 0.7389 & 0.6241 \\
     Baseline & 91.4947 & 0.6722 & 0.3917 \\
     \hline
     
\end{tabular}
\end{table}

\begin{figure}
    \centering
    \includegraphics[width=\textwidth]{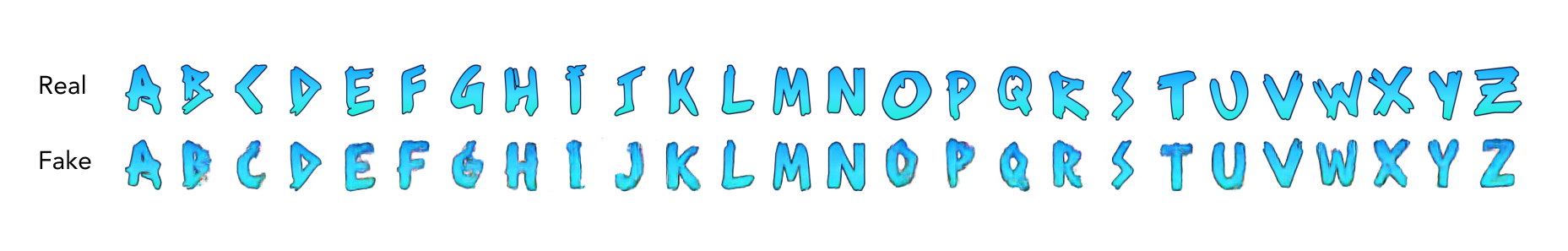}
    \caption{\label{fig:example}Example of one set of generated font}
\end{figure}

From the $L_1$ losses computed during fine-tuning shown in Figure \ref{fig:l1loss}, we observed that our baseline model got almost the same performance as stated in the original paper\cite{AGIS}. We also compare the performance of the baseline model with AGIS-Net (setting $m$=7, $n$=2) using the evaluation metrics used in their paper which are FID, SSIM, and pixel-level accuracy. Higher values of SSIM and pix-acc are better, whereas for FID, the lower the better. We didn't measure the inception score of our results since it can not directly reflect the quality of our synthesized glyph images. As shown in Table \ref{table:eval}, we achieved roughly the same performance in terms of SSIM. Regarding FID, the results are close, but for pix-acc we still need to improve our baseline model to match the ideal performance. An example of a set of generated fonts is shown in Figure \ref{fig:example}.

\section{Extended Methods}

AGIS-Net considers font generation as an image-to-image translation task. The difference is that it takes two conditions, which are style and content images. However, the original paper only states that it is capable to generate fonts using styles from the same language. Moreover, it heavily relies on the fine-tuning process. To extend it, we leveraged the idea of layer attention and context-aware attention as shown in Figure \ref{fig:layerattention} and Figure \ref{fig:contextattention} modules in the generator of Font Translator GAN. The context-aware attention network captures the local style feature, while the layer attention network captures the global style feature. They work together to make our generator more flexible when dealing with arbitrary font styles.

To be more detailed, in the context-aware attention network, we introduce a self-attention module to make the new feature contain both the information in their receptive fields and contextual information from other locations. We followed the suggestions in the original paper to use 3 parallel context-aware attentions.

After having local features and global features, we apply layer attention to select which feature level the model should focus on by assigning scores to them. After that, we take the mean and expand the encoded style features.

\begin{figure}
    \centering
    \includegraphics[width=1.0\textwidth]{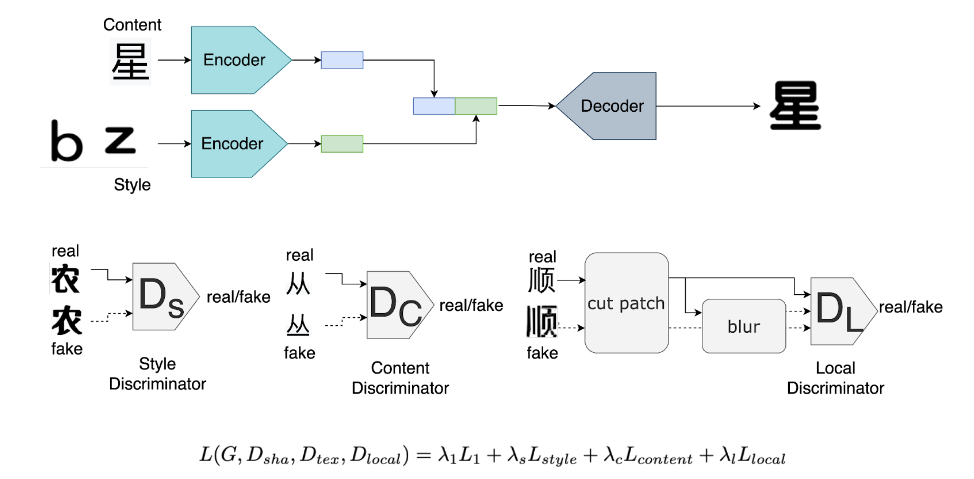}
    \caption{\label{fig:gasnext}GAS-NeXt architecture}
\end{figure}

For discriminators, instead of using only context and style discriminators as in Font Translator GAN, we keep using the local discriminator to retain a better local structure performance. We also adapt the objective function and adversarial loss of AGIS-Net. Overall, we show our architecture in Figure \ref{fig:gasnext}.

\section{Results and Discussions}

In our experiments with GAS-NeXt, we chose to use the dataset proposed by Li et al. \cite{ftransgan} to compare the cross-lingual font generation effects against the state-of-the-art model Font Translator GAN. The metrics comparison is shown in Table \ref{table:gasnextvsftransgan}. The metrics of GAS-NeXt are very similar to that of Font Translator GAN. 

\begin{figure}
    \centering
    \includegraphics[width=\textwidth]{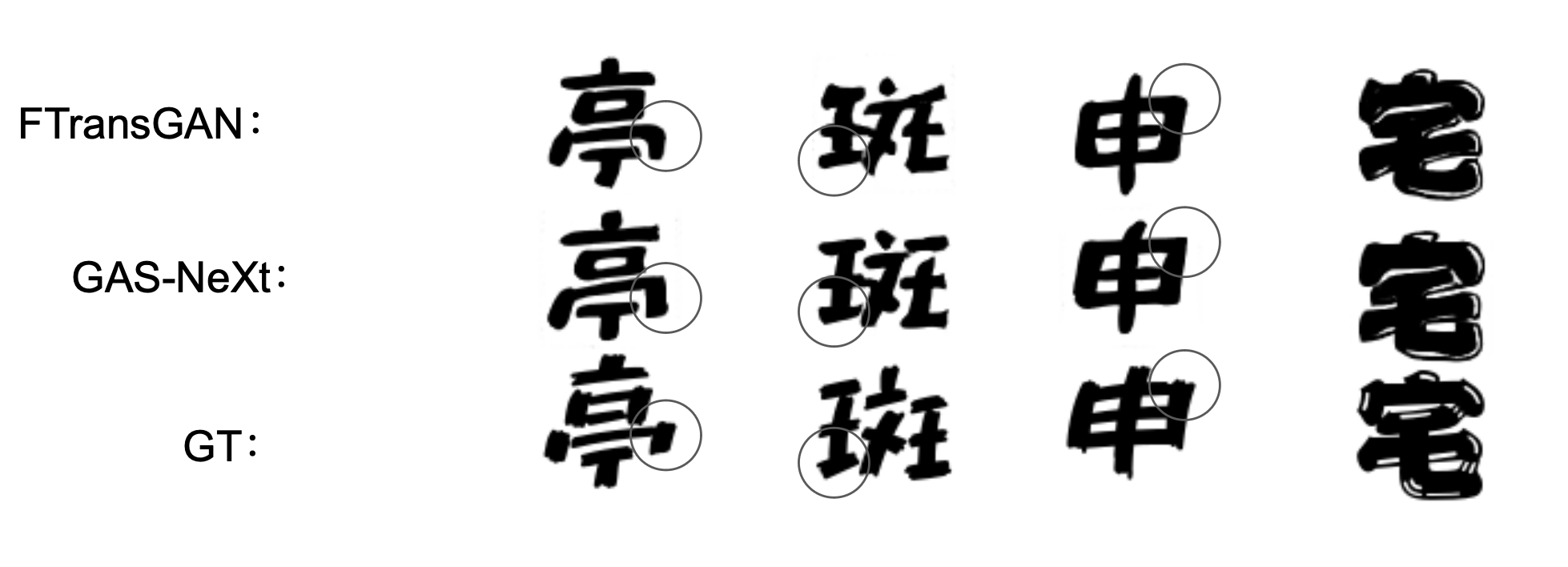}
    \caption{\label{fig:visualcomp}Visual results comparison between FTransGAN and GAS-NeXt}
\end{figure}

However, when examining the generation results, we can see that GAS-NeXt achieves a similar level of visual results as Font Translator GAN for fonts with similar local attributes as conventional fonts, while GAS-NeXt is able to achieve a better visual result for fonts with distinct local attributes. As shown in Figure \ref{fig:visualcomp}, we can clearly observe a better feature captured by GAS-NeXt in the circled regions. These regions have the commonality of being sharp-edged in the target font but are more rounded in conventional Chinese writings. This visual observation suggests that GAS-NeXt is able to capture local attributes better than FTransGAN, which is a crucial factor in generating a font with distinct styles. 

This improvement in learning local distinct styles can be attributed to our addition of the local discriminator inspired by AGIS-Net. By generating more fake samples from a real sample's local regions using patch cutting and blurring, and using a third discriminator specifically learning the distinctions in the local features, our model can capture these local distinctions and generate results more similar to the ground truth. 

\textbf{\begin{table}[]
\caption{\label{table:gasnextvsftransgan} Evaluation Metrics of Font Translator GAN and GAS-NeXt}
\centering
\begin{tabular}{@{}lllll@{}}
\toprule
\textbf{Content / Style} & \textbf{Accuracy} & \textbf{L1} & \textbf{SSIM} & \textbf{FID} \\
\midrule
\begin{tabular}[c]{@{}l@{}}FTransGAN\\ Unknown Content\end{tabular} & \begin{tabular}[c]{@{}l@{}}97.28 (Content)\\ 55.59 (Style)\end{tabular} & \begin{tabular}[c]{@{}l@{}}0.12\\ 0.12\end{tabular}     & \begin{tabular}[c]{@{}l@{}}0.50\\ 0.49\end{tabular}     & \begin{tabular}[c]{@{}l@{}}50.44\\ 321.60\end{tabular} \\   \\
\begin{tabular}[c]{@{}l@{}}FTransGAN\\ Unknown Style\end{tabular}   & \begin{tabular}[c]{@{}l@{}}99.88\\ 9.65\end{tabular}                      & \begin{tabular}[c]{@{}l@{}}0.18\\ 0.18\end{tabular}     & \begin{tabular}[c]{@{}l@{}}0.37\\ 0.37\end{tabular}     & \begin{tabular}[c]{@{}l@{}}96.79\\ 436.75\end{tabular}  \\  \\
\textbf{\begin{tabular}[c]{@{}l@{}}GasNeXt\\ Unknown Content\end{tabular}}     & \begin{tabular}[c]{@{}l@{}}96.52\\ 55.01\end{tabular}                     & \begin{tabular}[c]{@{}l@{}}0.1241\\ 0.1241\end{tabular} & \begin{tabular}[c]{@{}l@{}}0.4928\\ 0.4928\end{tabular} & \begin{tabular}[c]{@{}l@{}}52.012\\ 324.33\end{tabular} \\  \\
\textbf{\begin{tabular}[c]{@{}l@{}}GasNeXt\\ Unknown Style\end{tabular}}       & \begin{tabular}[c]{@{}l@{}}99.755\\ 9.05\end{tabular}                     & \begin{tabular}[c]{@{}l@{}}0.1865\\ 0.1865\end{tabular} & \begin{tabular}[c]{@{}l@{}}0.3459\\ 0.3459\end{tabular} & \begin{tabular}[c]{@{}l@{}}103.933\\ 460.827\end{tabular} \\ \bottomrule
\end{tabular}
\end{table}
}

\section{Conclusion and Future Work}

In this paper, we proposed a novel few-shot cross-lingual font generator network GAS-NeXt. Using the layer attention and context-aware attention methods from Font Translator GAN, and the idea of a local discriminator from AGIS-Net, we are able to achieve a similar metric with the state-of-the-art model Font Translator GAN, while obtaining better visual results in fonts with distinct local features. Although previous methods such as character decomposition also attempt to address this problem, these methods are not able to transfer the decomposition knowledge of characters from one language to another. To our knowledge, there have not been literature discussions focusing on cross-lingual font generation featuring local features. Our model lays a groundwork for future studies in improving learning cross-lingual font generation for fonts with distinct local features compared to conventional writing.

In addition to further experimentation on the local discriminator, another potential future work is more experimentation on colored font styles. Currently, the colored font style generation with cross-lingual capability for our model is non-ideal. We experimented on separating the three channels of image inputs and experiment on each channel independently, then reconstructed the three generated channels into a single image. However, this performed poorly as the three channels generated results in different pixel locations, causing incomplete overlaps.

%————————————————————————————————————————————————————————————————————————————————————

\pagebreak

\medskip
{\small
\bibliographystyle{plain}
\bibliography{references.bib}

\begin{thebibliography}{10}

\bibitem{azadi}
Samaneh Azadi, Matthew Fisher, Vladimir Kim, Zhaowen Wang, Eli Shechtman, and
  Trevor Darrell.
\newblock Multi-content gan for few-shot font style transfer, 2017.

\bibitem{interpolation}
Neill D.~F. Campbell and Jan Kautz.
\newblock Learning a manifold of fonts.
\newblock {\em ACM Trans. Graph.}, 33(4), jul 2014.

\bibitem{DMFont}
Junbum Cha, Sanghyuk Chun, Gayoung Lee, Bado Lee, Seonghyeon Kim, and Hwalsuk
  Lee.
\newblock Few-shot compositional font generation with dual memory, 2020.

\bibitem{figr}
Louis Clouâtre and Marc Demers.
\newblock Figr: Few-shot image generation with reptile, 2019.

\bibitem{gatys2016image}
Leon~A Gatys, Alexander~S Ecker, and Matthias Bethge.
\newblock Image style transfer using convolutional neural networks.
\newblock In {\em Proceedings of the IEEE conference on computer vision and
  pattern recognition}, pages 2414--2423, 2016.

\bibitem{Ghojogh_2020}
Benyamin Ghojogh, Fakhri Karray, and Mark Crowley.
\newblock Theoretical insights into the use of structural similarity index in
  generative models and inferential autoencoders.
\newblock In {\em Lecture Notes in Computer Science}, pages 112--117. Springer
  International Publishing, 2020.

\bibitem{GAN}
Ian Goodfellow, Jean Pouget-Abadie, Mehdi Mirza, Bing Xu, David Warde-Farley,
  Sherjil Ozair, Aaron Courville, and Yoshua Bengio.
\newblock Generative adversarial nets.
\newblock In Z.~Ghahramani, M.~Welling, C.~Cortes, N.~Lawrence, and K.Q.
  Weinberger, editors, {\em Advances in Neural Information Processing Systems},
  volume~27. Curran Associates, Inc., 2014.

\bibitem{GlyphGAN}
Hideaki Hayashi, Kohtaro Abe, and Seiichi Uchida.
\newblock Glyphgan: Style-consistent font generation based on generative
  adversarial networks.
\newblock {\em CoRR}, abs/1905.12502, 2019.

\bibitem{huang2017real}
Haozhi Huang, Hao Wang, Wenhan Luo, Lin Ma, Wenhao Jiang, Xiaolong Zhu, Zhifeng
  Li, and Wei Liu.
\newblock Real-time neural style transfer for videos.
\newblock In {\em Proceedings of the IEEE Conference on Computer Vision and
  Pattern Recognition}, pages 783--791, 2017.

\bibitem{huang2017arbitrary}
Xun Huang and Serge Belongie.
\newblock Arbitrary style transfer in real-time with adaptive instance
  normalization.
\newblock In {\em Proceedings of the IEEE international conference on computer
  vision}, pages 1501--1510, 2017.

\bibitem{RDGAN}
Yaoxiong Huang, Mengchao He, Lianwen Jin, and Yongpan Wang.
\newblock Rd-gan: Few/zero-shot chinese character style transfer via radical
  decomposition and rendering.
\newblock In Andrea Vedaldi, Horst Bischof, Thomas Brox, and Jan-Michael Frahm,
  editors, {\em Computer Vision -- ECCV 2020}, pages 156--172, Cham, 2020.
  Springer International Publishing.

\bibitem{pix2pix}
Phillip Isola, Jun-Yan Zhu, Tinghui Zhou, and Alexei~A. Efros.
\newblock Image-to-image translation with conditional adversarial networks,
  2016.

\bibitem{CGGAN}
Yuxin Kong, Canjie Luo, Weihong Ma, Qiyuan Zhu, Shenggao Zhu, Nicholas Yuan,
  and Lianwen Jin.
\newblock Look closer to supervise better: One-shot font generation via
  component-based discriminator, 2022.

\bibitem{DRIT}
Hsin-Ying Lee, Hung-Yu Tseng, Jia-Bin Huang, Maneesh~Kumar Singh, and
  Ming-Hsuan Yang.
\newblock Diverse image-to-image translation via disentangled representations.
\newblock In {\em European Conference on Computer Vision}, 2018.

\bibitem{ftransgan}
Chenhao Li, Yuta Taniguchi, Min Lu, and Shin'ichi Konomi.
\newblock Few-shot font style transfer between different languages.
\newblock In {\em Proceedings of the IEEE/CVF Winter Conference on Applications
  of Computer Vision}, pages 433--442, 2021.

\bibitem{li2017universal}
Yijun Li, Chen Fang, Jimei Yang, Zhaowen Wang, Xin Lu, and Ming-Hsuan Yang.
\newblock Universal style transfer via feature transforms.
\newblock {\em Advances in neural information processing systems}, 30, 2017.

\bibitem{casia}
Cheng-Lin Liu, Fei Yin, Da-Han Wang, and Qiu-Feng Wang.
\newblock Casia online and offline chinese handwriting databases.
\newblock In {\em 2011 International Conference on Document Analysis and
  Recognition}, pages 37--41, 2011.

\bibitem{funit}
Ming-Yu Liu, Xun Huang, Arun Mallya, Tero Karras, Timo Aila, Jaakko Lehtinen,
  and Jan Kautz.
\newblock Few-shot unsupervised image-to-image translation.
\newblock 2019.

\bibitem{context_loss}
Roey Mechrez, Itamar Talmi, and Lihi Zelnik{-}Manor.
\newblock The contextual loss for image transformation with non-aligned data.
\newblock {\em CoRR}, abs/1803.02077, 2018.

\bibitem{cGAN}
Mehdi Mirza and Simon Osindero.
\newblock Conditional generative adversarial nets, 2014.

\bibitem{mcgan}
Hyojin Park, YoungJoon Yoo, and Nojun Kwak.
\newblock Mc-gan: Multi-conditional generative adversarial network for image
  synthesis, 2018.

\bibitem{CCFont}
Jangkyoung Park, Ammar~Ul Hassan, and Jaeyoung Choi.
\newblock Ccfont: Component-based chinese font generation model using
  generative adversarial networks (gans).
\newblock {\em Applied Sciences}, 12(16), 2022.

\bibitem{MXFont}
Song Park, Sanghyuk Chun, Junbum Cha, Bado Lee, and Hyunjung Shim.
\newblock Multiple heads are better than one: Few-shot font generation with
  multiple localized experts, 2021.

\bibitem{zi2zi}
Yuchen Tian.
\newblock zi2zi: Master chinese calligraphy with conditional adversarial
  networks, 2017.

\bibitem{tuyen2021deep}
Nguyen~Quang Tuyen, Son~Truong Nguyen, Tae~Jong Choi, and Vinh~Quang Dinh.
\newblock Deep correlation multimodal neural style transfer.
\newblock {\em IEEE Access}, 9:141329--141338, 2021.

\bibitem{verma2018neural}
Prateek Verma and Julius~O Smith.
\newblock Neural style transfer for audio spectograms.
\newblock {\em arXiv preprint arXiv:1801.01589}, 2018.

\bibitem{4775883}
Zhou Wang and Alan~C. Bovik.
\newblock Mean squared error: Love it or leave it? a new look at signal
  fidelity measures.
\newblock {\em IEEE Signal Processing Magazine}, 26(1):98--117, 2009.

\bibitem{CalliGAN}
Shan-Jean Wu, Chih-Yuan Yang, and Jane Yung-jen Hsu.
\newblock Calligan: Style and structure-aware chinese calligraphy character
  generator, 2020.

\bibitem{Yang2019TETGAN}
Shuai Yang, Jiaying Liu, Wenjing Wang, and Zongming Guo.
\newblock Tet-gan: Text effects transfer via stylization and destylization.
\newblock In {\em AAAI Conference on Artificial Intelligence}, 2019.

\bibitem{AGIS}
Gao Yue, Guo Yuan, Lian Zhouhui, Tang Yingmin, and Xiao Jianguo.
\newblock Artistic glyph image synthesis via one-stage few-shot learning.
\newblock {\em ACM Trans. Graph.}, 38(6), 2019.

\bibitem{storkegan}
Jinshan Zeng, Qi~Chen, Yunxin Liu, Mingwen Wang, and Yuan Yao.
\newblock Strokegan: Reducing mode collapse in chinese font generation via
  stroke encoding.
\newblock 2020.

\bibitem{zhang2018separating}
Yexun Zhang, Ya~Zhang, and Wenbin Cai.
\newblock Separating style and content for generalized style transfer.
\newblock In {\em Proceedings of the IEEE conference on computer vision and
  pattern recognition}, pages 8447--8455, 2018.

\bibitem{cyclegan_zhu}
Jun-Yan Zhu, Taesung Park, Phillip Isola, and Alexei~A. Efros.
\newblock Unpaired image-to-image translation using cycle-consistent
  adversarial networks, 2017.

\bibitem{zhu2017}
Jun-Yan Zhu, Richard Zhang, Deepak Pathak, Trevor Darrell, Alexei~A. Efros,
  Oliver Wang, and Eli Shechtman.
\newblock Toward multimodal image-to-image translation, 2017.

\end{thebibliography}
}
\raggedbottom
\pagebreak

\appendix
\section{AGIS-Net}
\begin{figure}[!htb]
    \centering
    \includegraphics[width=1.0\textwidth]{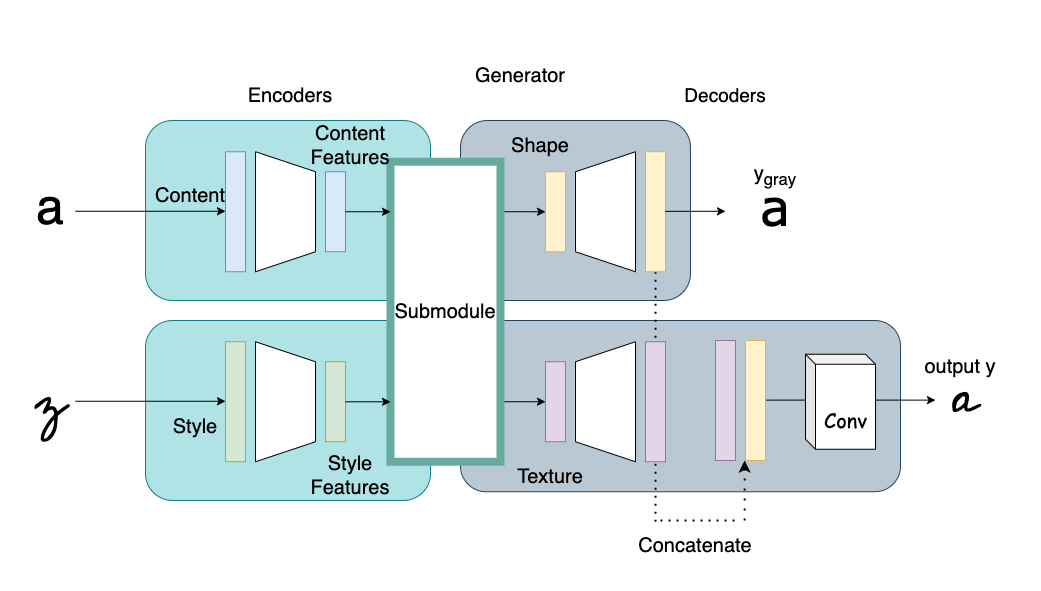}
    \caption{\label{fig:baseline_generator}AGIS-Net Generator}
\end{figure}

\begin{figure}[!htb]
    \centering
    \includegraphics[width=1.0\textwidth]{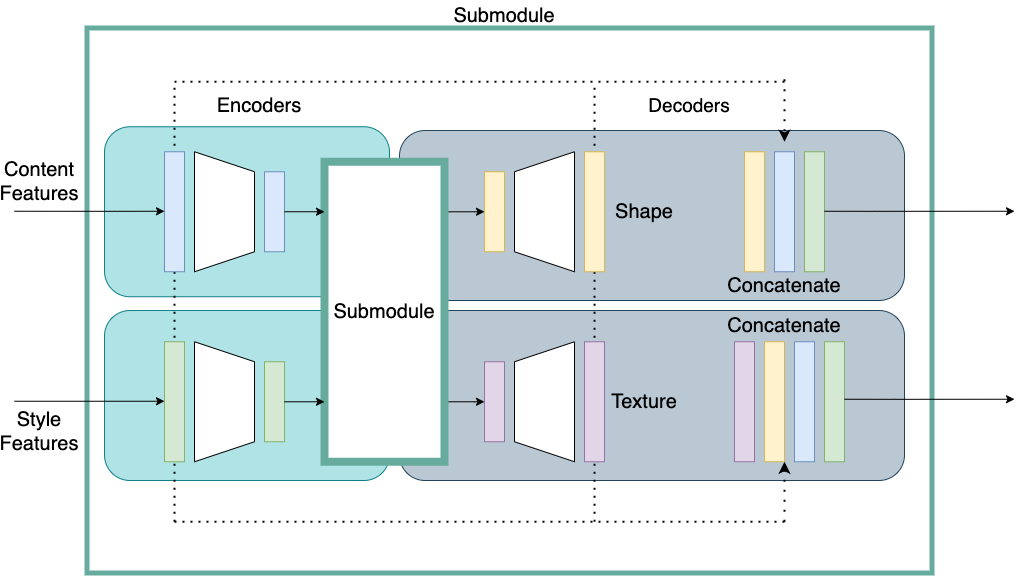}
    \caption{\label{fig:baseline_submodule}AGIS-Net Generator Submodule}
\end{figure}
\pagebreak

\section{Metrics for Baseline}
\begin{figure}[!htb]
    \centering
    \includegraphics[width=1.0\textwidth]{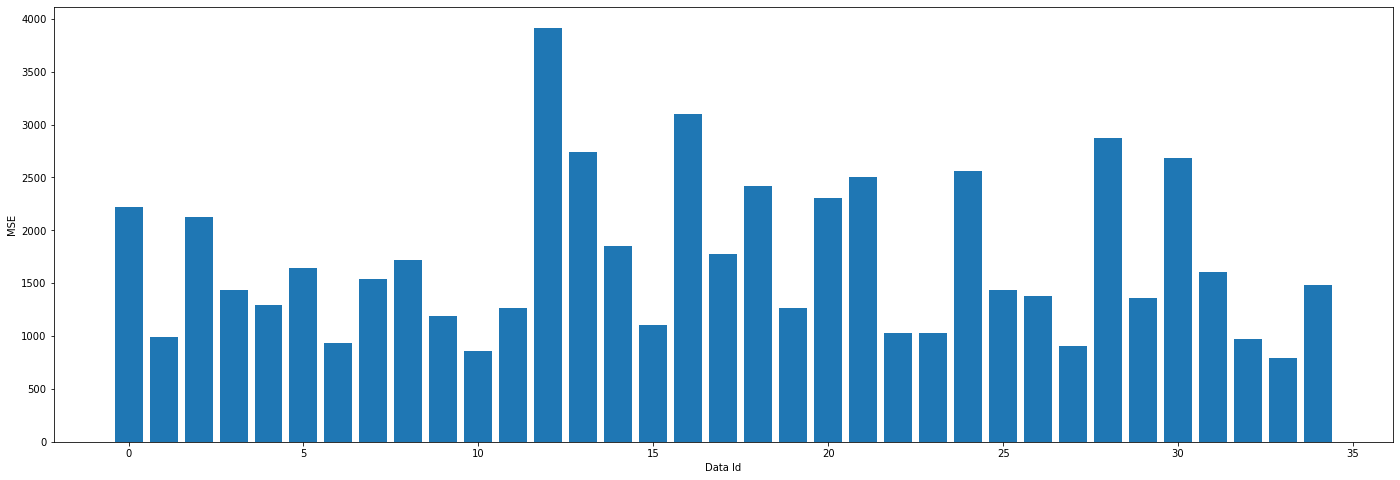}
    \caption{\label{fig:mse}MSE of test dataset}
\end{figure}

\begin{figure}[!htb]
    \centering
    \includegraphics[width=1.0\textwidth]{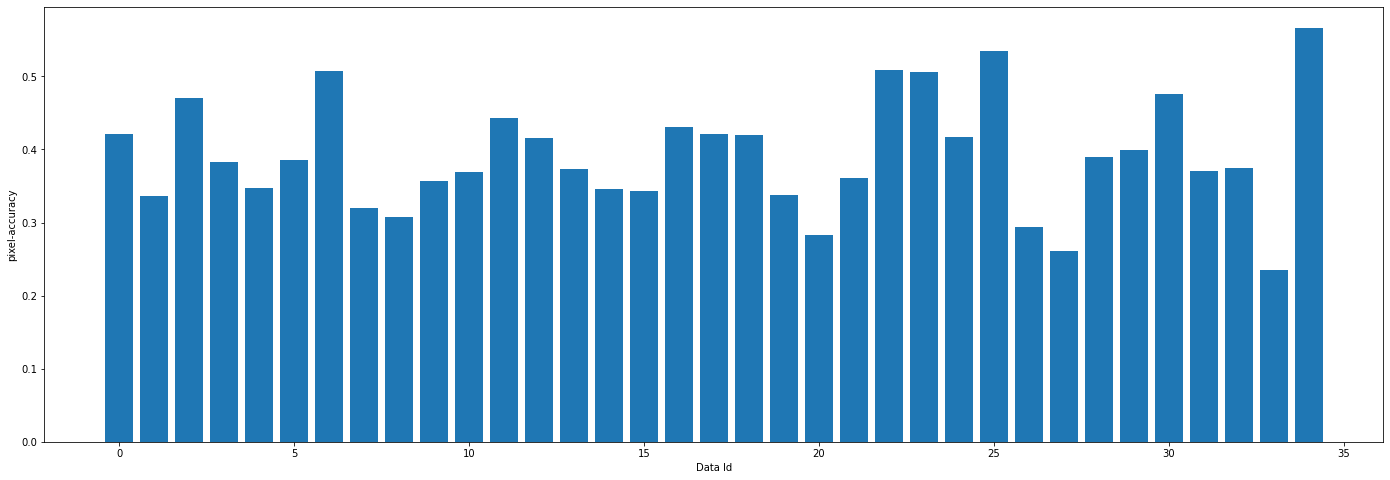}
    \caption{\label{fig:pixacc}pixel-level accuracy of test dataset}
\end{figure}

\begin{figure}[!htb]
    \centering
    \includegraphics[width=1.0\textwidth]{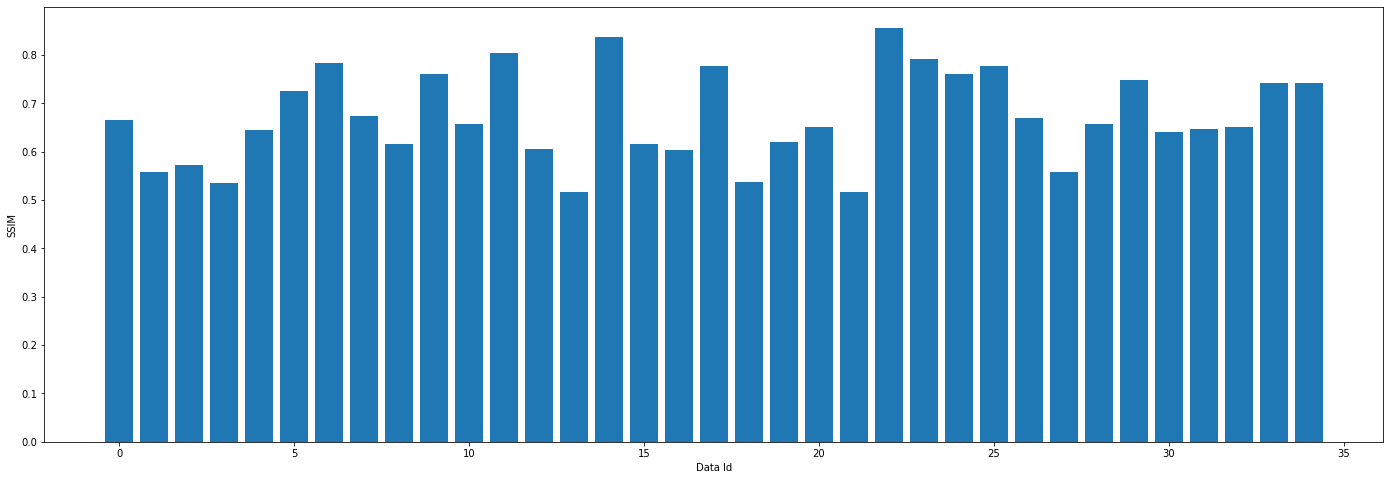}
    \caption{\label{fig:ssim}SSIM of test dataset}
\end{figure}
\pagebreak

\section{L1 Loss for baseline model}
\begin{figure}[!htb]
    \centering
    \includegraphics[width=1.0\textwidth]{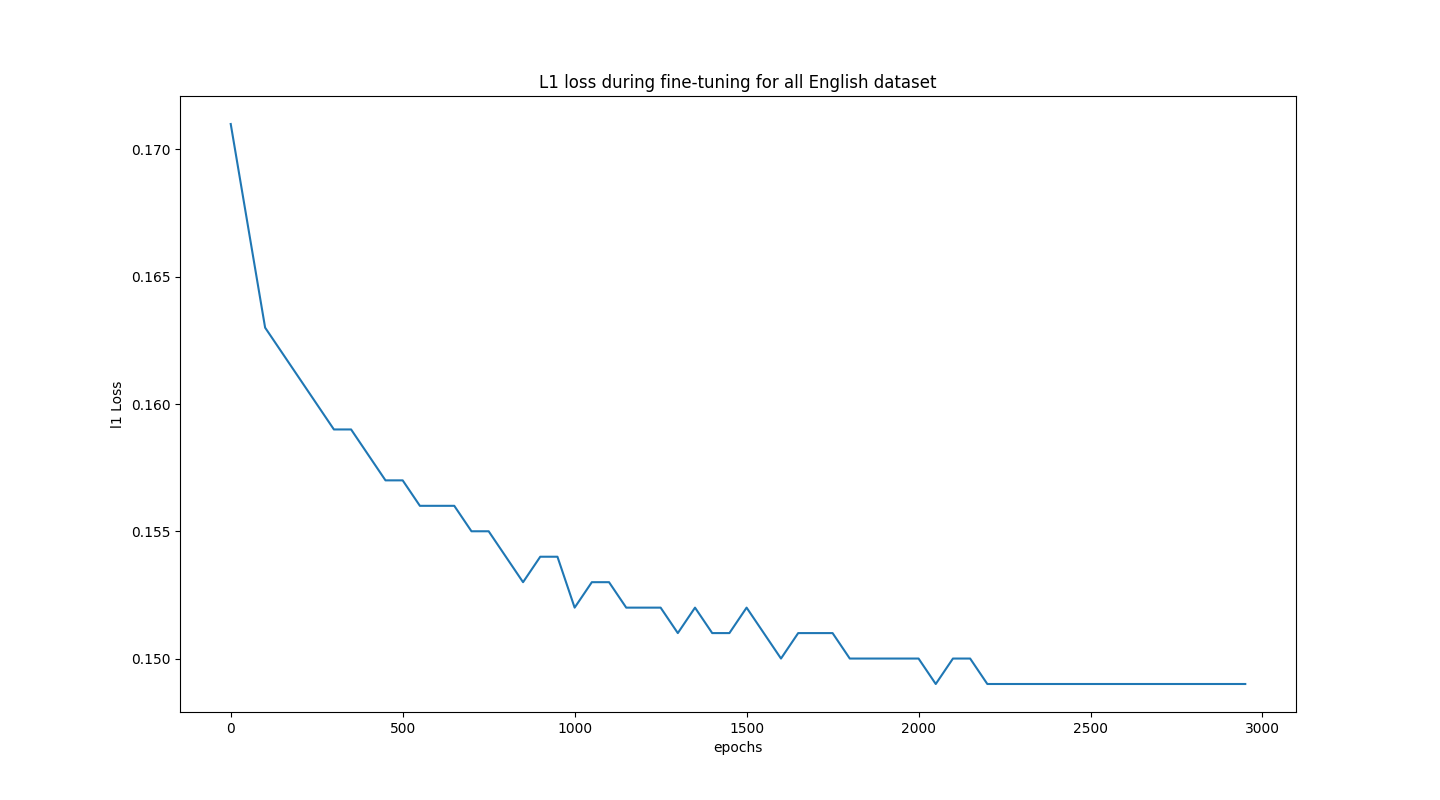}
    \caption{\label{fig:l1loss}L1 Loss in Fine-Tuning Stage}
\end{figure}
\pagebreak

\section{Context-aware Attention}
\begin{figure}[!htb]
    \includegraphics[width=1.0\textwidth]{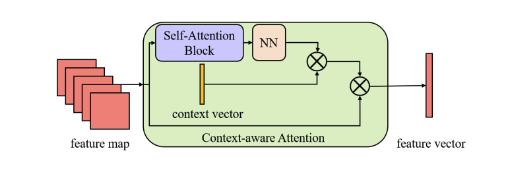}
    \caption{\label{fig:contextattention}Context-aware attention}
\end{figure}
\pagebreak

\section{Layer Attention}
\begin{figure}[!htb]
    \centering
    \includegraphics[width=1.0\textwidth]{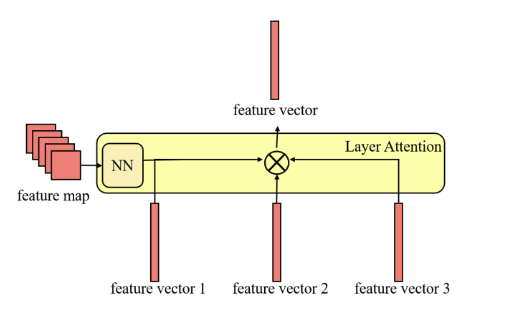}
    \caption{\label{fig:layerattention}Layer attention}
\end{figure}
\pagebreak

\end{document}